\documentclass{article}

\usepackage[preprint]{neurips_2025}

\usepackage[utf8]{inputenc} 
\usepackage[T1]{fontenc}    
\usepackage{hyperref}       
\usepackage{url}            
\usepackage{booktabs}       
\usepackage{amsfonts}       
\usepackage{amsmath}        
\usepackage{amssymb}        
\usepackage{nicefrac}       
\usepackage{microtype}      
\usepackage{multirow}       
\usepackage{graphicx}
\usepackage{enumitem}
\usepackage{caption}
\usepackage{subcaption}
\usepackage{wrapfig}     
\usepackage{booktabs}    
\usepackage[table,xcdraw]{xcolor}
\newtheorem{theorem}{Theorem}[section]

\usepackage{colortbl}

\definecolor{Col1}{HTML}{F0F3FF}  
\definecolor{Col2}{HTML}{FFF5E6}  
\definecolor{Col3}{HTML}{F0FFE6}  

\title{
    \includegraphics[width=0.65cm]{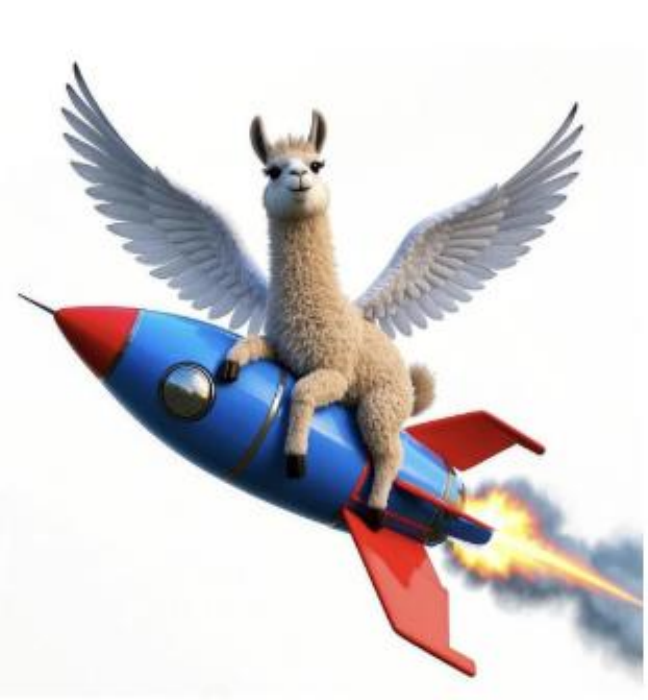}
    Scaling Laws for Speculative Decoding
}

\author{%
  Siyuan Yan†$^{1,3}$, Mo Zhu†$^{1,4}$, Guo-qing Jiang$^{1}$\thanks{Corresponding Author, † Equal Contribution}, Jianfei Wang$^{1}$, Jiaxing Chen$^{1}$, Wentai Zhang$^{1}$,\\ 
  \textbf{Xiang Liao$^{1,2}$, Xiao Cui$^{5}$, Chen Zhang$^{2}$, Zhuoran Song$^{2}$, Ran Zhu$^{1}$} \\
  $^{1}$Red Note Hi-Lab, $^{2}$Shanghai Jiaotong University, $^{3}$Nanjing University, \\
  $^{4}$Zhejiang University, $^{5}$Chinese University of Hong Kong \\
  \texttt{\{yansiyuan,zhumo,liwendao,wangjianfei,chenjiaxing,zhangwentai,zhuran\}} \\
  \texttt{@xiaohongshu.com, \{liaoxiang,chenzhang.sjtu,songzhuoran\}@sjtu.edu.cn,} \\
  \texttt{cuixiao2001@mail.ustc.edu.cn}
}

\newcommand{\seqpre}{s_{\text{pre}}}
\newcommand{\hkv}{h_{\text{kv}}}
\newcommand{\hmlp}{h_{\text{mlp}}}
\newcommand{\topk}{\text{top}_k}
\newcommand{\tacc}{t_{\text{acc}}}

\begin{document}

\maketitle

\begin{abstract}

The escalating demand for efficient decoding in large language models (LLMs) is particularly critical for reasoning-intensive architectures like OpenAI-o3 and DeepSeek-R1, which depend on extended \textit{chain-of-thought} reasoning. This study investigates speculative decoding techniques through dense LLM architectures to establish foundational insights for accelerating reasoning tasks. While speculative decoding methods leveraging parallel draft-verification cycles have emerged as promising acceleration techniques, the scaling laws governing decoding efficiency remain under-explored compared to conventional backbone LLMs developed through Pretraining$\to$SFT$\to$RLHF training paradigms.
In this work, we discover \textbf{Log-linear Scaling Laws} (Theorem~\ref{theorem:scaling_law_pt_tokens}, \ref{theorem:scaling_law_model_capacity} \& \ref{theorem:scaling_law_decoding_bs}) governing draft model acceptance rate (or decoding speed) across three dimensions: pretraining token volume, draft model capacity, and decoding batch size. Building on these laws, we achieve \textit{Scylla}, which coordinates multi-dimensional scaling for popular LLMs (Llama2/3, Qwen2.5).
Empirical validation shows Scylla achieves \textbf{1.5-2.2} higher acceptance rate than EAGLE2 and \textbf{0.3} higher than EAGLE3 at temperature \(T = 0\), with peak performance gains on summarization and QA tasks (Figure~\ref{fig1:best_result}). Industrial inference engine deployments demonstrate \textbf{2$\times$} decoding throughput improvements over EAGLE2 (Table~\ref{table:decoing_peak_throughput}), validating the transformative potential of systematic scaling for efficient LLM inference. \textcolor{blue}{\href{https://None}{Code}} will be released later.
\end{abstract}


\section{Introduction}

The scaling laws governing LLMs have been extensively investigated in prior works~\cite{kaplan2020scaling, hoffmann2022training}.These studies reveal robust power-law relationships between cross-entropy loss and key computational factors including model size, dataset size, and FLOPs. Recent advancements exemplified by models such as GPT4-o3 and DeepSeek-R1~\cite{guo2025deepseek} demonstrate that scaling up test-time computations can significantly push the boundaries of model capabilities, enabling human-level reasoning abilities. However, these advances typically require a lengthy thinking process during autoregressive decoding, resulting in high computational costs and latency. Memory-bandwidth constraints further exacerbate underutilization of compute resources during token generation. Speculative decoding methods, including Medusa~\cite{cai2024medusa} and EAGLE series~\cite{li2024eagle, li2024eagle2, li2025eagle3}, address these challenges by predicting multi-step tokens with minimal overhead and verifying them with underutilized computational resources, thereby significantly improving decoding throughput.

\begin{figure}[t]
  \centering
  \begin{subfigure}[b]{0.32\textwidth}
    \centering
    \includegraphics[width=\linewidth]{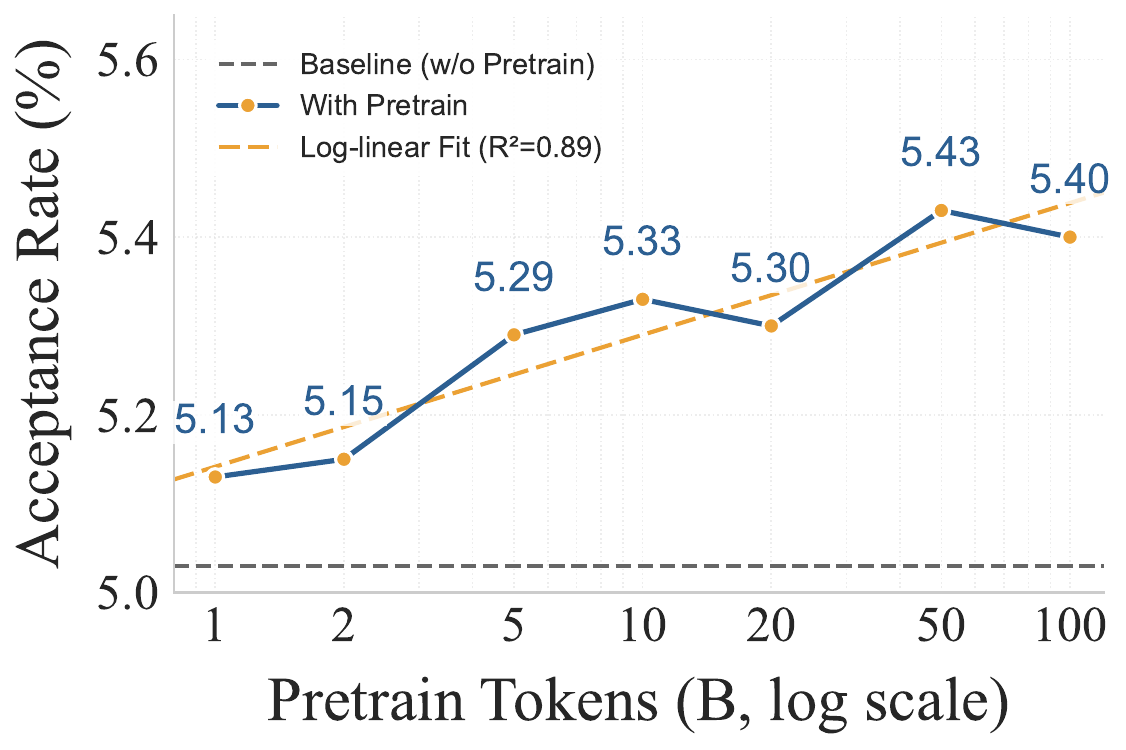}
    \label{fig:scaling_pretrain_tokens}
  \end{subfigure}\hfill
  \begin{subfigure}[b]{0.32\textwidth}
    \centering
    \includegraphics[width=\linewidth]{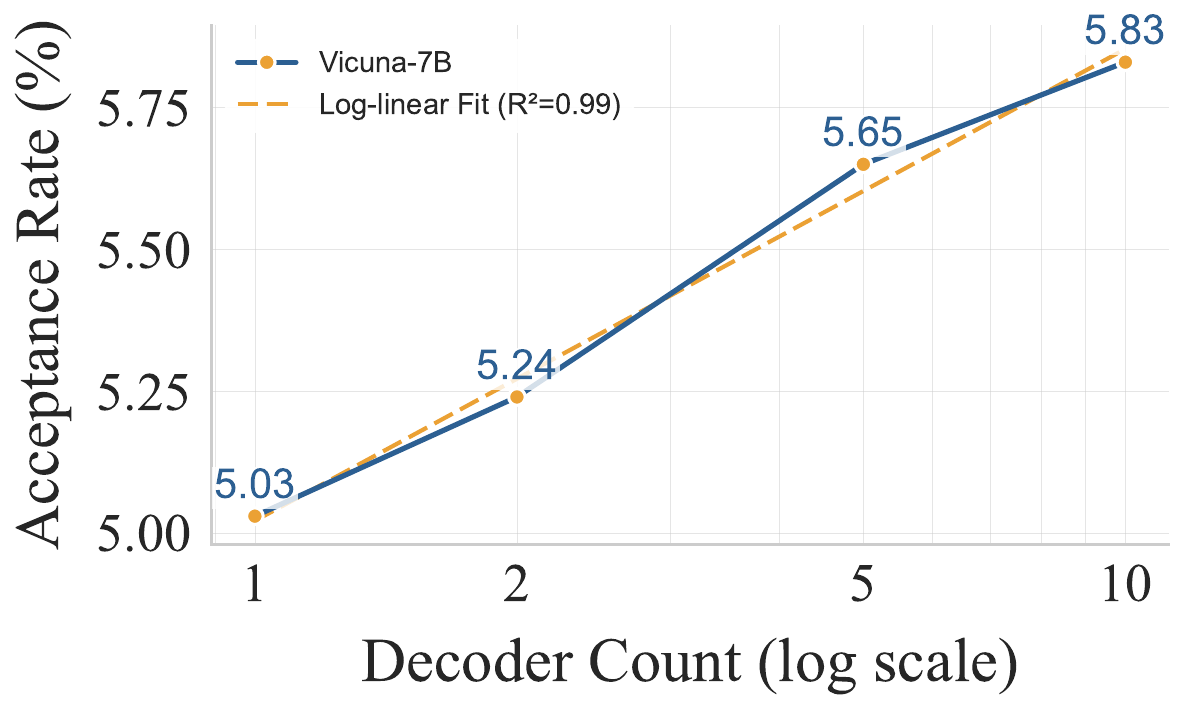}
    \label{fig:scaling_decoder_layers}
  \end{subfigure}\hfill
  \begin{subfigure}[b]{0.32\textwidth}
    \centering
    \includegraphics[width=\linewidth]{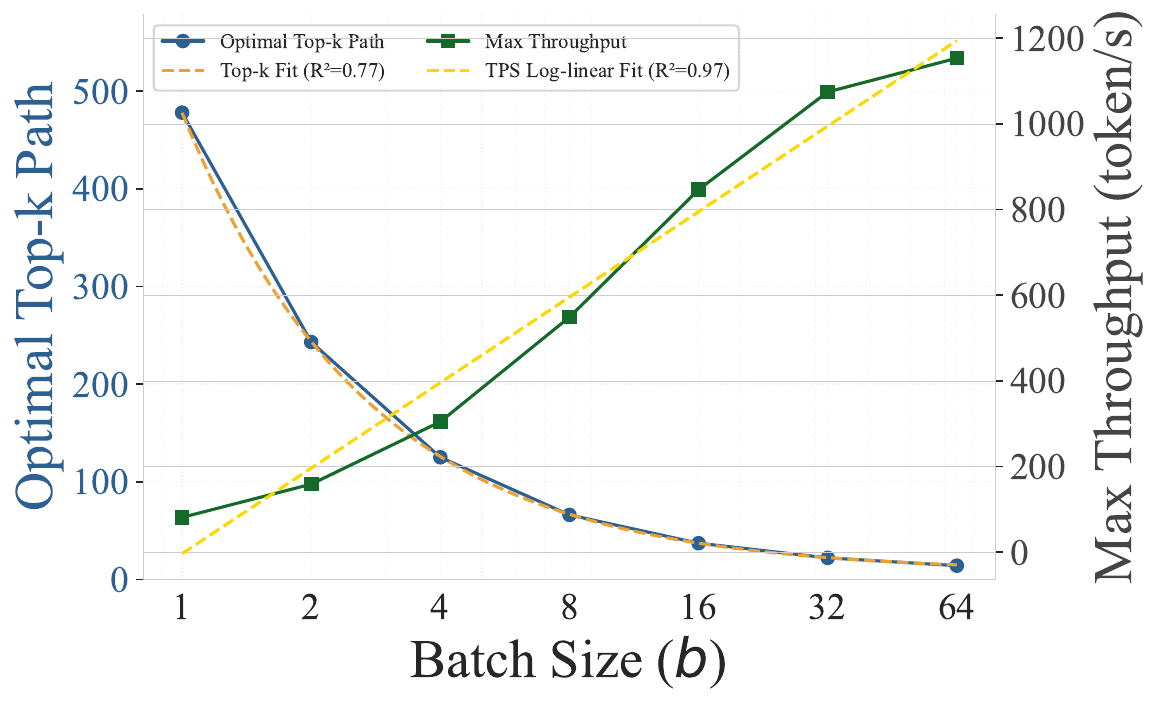}
    \label{fig:optimal_topk_vs_batch}
  \end{subfigure}

  \caption{(Left): Scaling law of acceptance rate on MT‑Bench and pretrain tokens; (Middle): Scaling up decoder layers; (Right): Optimal topk‑path and throughput versus decoding batch size.}
  \label{fig:scaling_experiments}
\end{figure}

While scaling laws for full-cycle training paradigms (Pretraining$\to$SFT$\to$RLHF) have been well-established, to the best of our knowledge, the scaling laws governing the decoding
speeds of speculative decoding remain under-explored. GPT serious \cite{ouyang2022training, achiam2023gpt} posits that pretraining represents a more knowledge-rich phase for Transformer-based models, implying potential benefits from computational scaling in draft model pretraining. However, existing work predominantly focuses on architectural modifications and sampling strategies for draft models (Section~\ref{sec:related_works}), with limited exploration of scaling laws under Pretraining$\to$SFT paradigms. Through systematic scaling of pretraining tokens and draft model capacity (Section~\ref{sec:Scylla}), we identify the log-linear scaling laws between acceptance rate, pretrain tokens (Theorem~\ref{theorem:scaling_law_pt_tokens}, Figure~\ref{fig:scaling_experiments} Left) and draft model capacity (Theorem~\ref{theorem:scaling_law_model_capacity}, Figure~\ref{fig:scaling_experiments} Middle).

Moreover, existing studies typically focus on single-batch decoding scenarios, whereas industrial deployments often decouple prefill and decoding phases while processing larger batches ($\text{batch} \gg 1$) for efficiency. Our theoretical analysis based on a roofline model reveals critical performance degradation patterns under scaling up batch sizes, for instance, Figure~\ref{fig:scaling_bs} shows throughput collapses when $\topk>16$ at $\text{batch}=64$, indicating the existence of batch-dependent scaling laws for optimal dynamic tree configurations and decoding speeds (Theorem~\ref{theorem:scaling_law_decoding_bs}, Figure~\ref{fig:scaling_experiments} Right panel). 

\begin{figure}
\begin{center}
\includegraphics[width=0.8\linewidth]{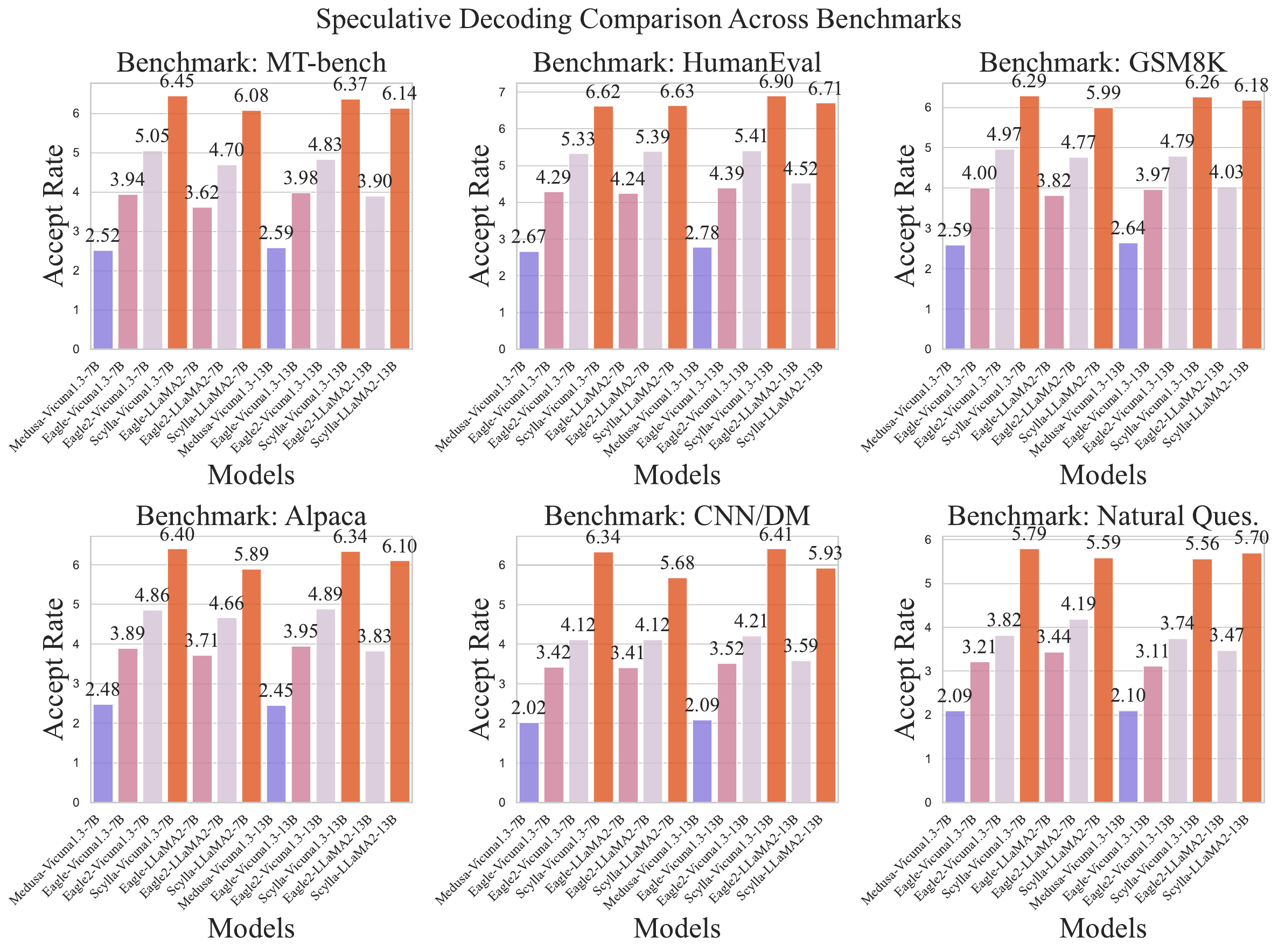}
\caption{Acceptance rate comparison between Scylla and previous SOTA methods across several benchmarks.}
\label{fig1:best_result}
\end{center}
\end{figure}

By leveraging these scaling laws through scaling up pretrain data, draft model capacity and decoding batch size, we develop Scylla shown in Figure~\ref{architecture} and it does well in many tasks as shown in Figure~\ref{fig1:best_result}. Our implementation achieves 1.5-2.2 higher acceptance rate than EAGLE2 at temperature \(T = 0\), with peak gains on summarization/QA tasks. After scaling up decoding batch size with the optimal $\topk$, Scylla achieves a 2$\times$ decoding speed over EAGLE2 on our industrial-level inference engine. Scylla uses only the 8× SFT data augmentation from EAGLE3 (excluding hidden fusion and train-time test features) and achieves a 0.3 higher acceptance rate than EAGLE3, demonstrating the effectiveness of our independent scaling approach (Appendix~\ref{app:eagle3}). The impact of RLHF on draft accuracy presents an intriguing open question for subsequent investigation.

\textbf{Summary of Scaling Laws.} 
The acceptance rate and decoding speed of speculative decoding with inference engine can be predicted by the following log-linear scaling laws, which are validated across more than ten experimental runs with consistent results.
\begin{theorem}[Log-linear Scaling Law of Pretrain tokens]
\label{theorem:scaling_law_pt_tokens}
\begin{equation}
\text{Acceptance rate} = \alpha \cdot \log_{10}(T_{\text{pretrain}}) + \beta; \quad\alpha \sim 0.08, \quad\beta \sim 5.05
\label{eq:acc_vs_t_pretrain}
\end{equation}
where $T_{\text{pretrain}}$ represents the number of pretrain tokens.
\end{theorem}

\begin{theorem}[Log-linear Scaling Law of Draft Model Capacity]
\label{theorem:scaling_law_model_capacity}
\begin{equation}
\text{Acceptance rate} = \alpha \cdot \log_{10}(D) + \beta; \quad\alpha \sim 0.74, \quad\beta \sim 4.61
\label{eq:acc_vs_D}
\end{equation}
where $D$ represents the count of draft model decoders.
\end{theorem}

\begin{theorem}[Log-linear Scaling Law of Decoding Batch Size]
\label{theorem:scaling_law_decoding_bs}
\begin{equation}
    \label{eq:tps_vs_b}
    \text{Throughtput} = \alpha \cdot \log_{2}(b) + \beta; \quad\alpha \sim 286.79, \quad\beta \sim 7.54
\end{equation}
\begin{equation}
\label{eq:topk_vs_b}
\text{Optimal}~\topk(b) = 27904\sqrt{1 + \frac{0.034}{b}} - 27897
\end{equation}
where $b$ means batch size and $\topk$ means TopK-paths during verification stage.
\end{theorem}

\setlength{\textfloatsep}{11pt}

\section{Preliminaries}
\label{Preliminaries}
\begin{figure}[t]
\begin{center}
\includegraphics[width=0.8\linewidth]{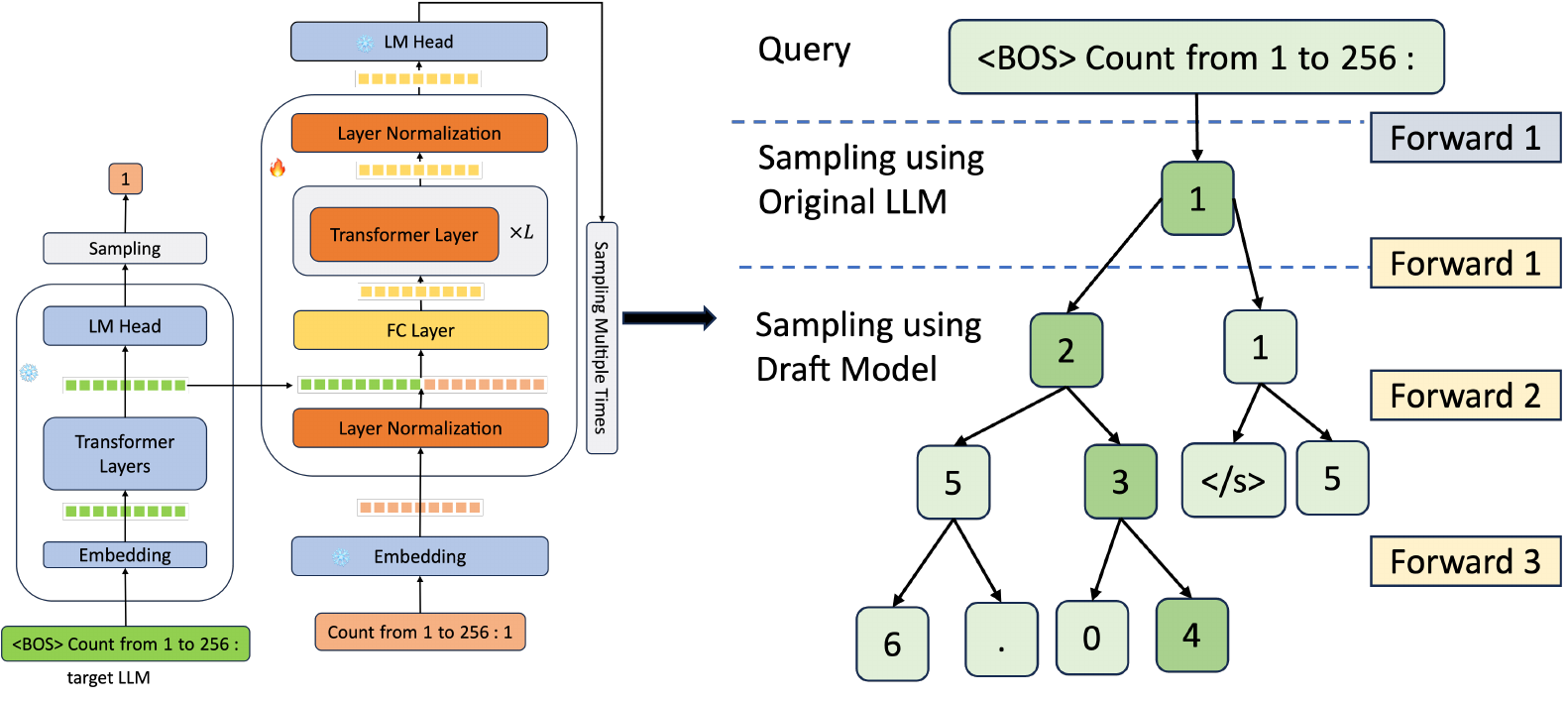}
\caption{Speculative decoding architecture used for scaling laws.}
\label{architecture}
\end{center}
\end{figure}

\textbf{Drafting and Verification.} Modern speculative decoding techniques~\cite{stern2018blockwise} employ a two-stage process: draft token generation and verification through a single forward pass of the backbone LLM. Let $T_{1:s}$ denote the prefix sequence containing $s$ confirmed tokens, and $h$ represent the hidden dimension size. During the drafting phase (Figure~\ref{architecture}), given input embeddings $E(T_{1:s}) \in \mathbb{R}^{s\times h}$ and backbone LLM hidden states $H \in \mathbb{R}^{s\times h}$, the draft model concats these two inputs at the last dimension ($h\to2h$) and performs dimension reduction through linear transformation $W_{FC} \in \mathbb{R}^{2h\times h}$ to obtain compressed representations $H' = HW_{FC}$. These representations are then fed through $L$ decoder layers and the backbone’s language modeling head. This generates draft token logits that align with the standard next-token prediction paradigm. The draft model autoregressively constructs a candidate tree $\tau = \{t_{s+1}^{(m)},\dots,t_{s+k_m}^{(m)}\}_{m=1}^M$ through multiple decoding steps ($m$ means the $m^{th}$ path of M). Each parent node $t_{s+k_m}^{(m)}$ spawns $top_c$ child nodes through conditional expansion: $\{t_{s+k_m+1}^{(m'_i)}\}_{i=1}^{top_c} \sim \hat{P}(\cdot | T_{1:s}, t_{s+1}^{(m)}, \dots, t_{s+k_m}^{(m)})$. This generates multiple candidate continuation paths $T_{s+1:s+k_m}^{(m)}$ for parallel verification. 

The verification phase employs the backbone LLM to compute target distribution $P(t_{s+1:s+k_m}^{(m)}|T_{1:s} \in \mathbb{R}^{k_m\times \left|V\right|})$ through masked parallel computation. For each candidate path $T_{s+1:s+k_m}^{(m)}$, the probability of sequential token acceptance is given by $\min(1, {P(t_{s+i}^{(m)})}/{\hat{P}(t_{s+i}^{(m)})})$. Upon rejection at position $i$, a replacement token is sampled from the distribution norm $\max(0, P(t_{s+i}| T_{1:s+i-1}) - \hat{P}(t_{s+i}|T_{1:s+i-1}))$, and subsequent candidate tokens are discarded. The verification efficiency derives from processing all candidate paths through optimized attention masking patterns (Figure~\ref{fig:attn_mask}), where the maximum verification depth corresponds to the longest accepted subsequence. This mechanism enables decoding $k$ tokens per cycle with computational complexity equivalent to a single backbone model forward pass, achieving substantial acceleration over vanilla auto-regressive decoding.

\textbf{Limitations.} Current research~\cite{li2024eagle, li2024eagle2, li2025eagle3} uses SFT-only training, deviating from the standard mainstream LLMs development pipeline (Pretraining$\to$SFT$\to$RLHF). 
While scaling pretraining data and model capacity has proven effective for enhancing backbone LLMs' performance, its impact on speculative decoding acceleration remains underexplored. Though speculative decoding leverages draft/backbone model speed gaps for efficient token generation, a critical question remains: memory-bound verification may not persist under large batch decoding.

\section{Methods}
\label{sec:Scylla}

This section introduces our scaling framework. We first establish the two-stage Pretraining→SFT pipeline with Transformer architecture, then investigate pretraining data scaling and draft model capacity expansion. We further develop a hardware-aware roofline model to analyze batch-size effects on arithmetic intensity and throughput, and ultimately propose an adaptive TopK-path selection strategy that maximizes decoding performance across varying batch sizes.

\subsection{Training Paradigm and Computational Scaling}
\begin{figure}[t]
    \centering
    \begin{subfigure}[b]{0.65\linewidth}
        \centering
        \includegraphics[width=\linewidth]{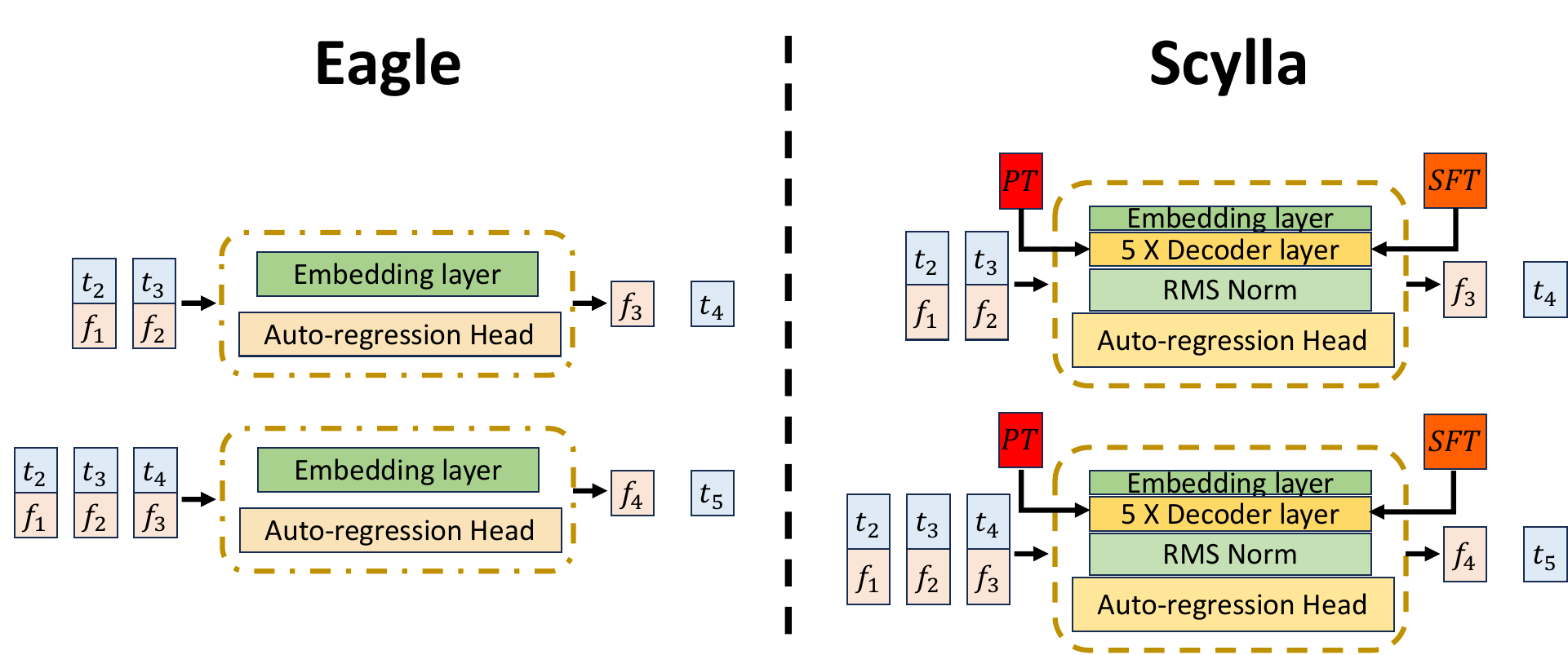}
        \caption{}
        \label{fig:architecture}
    \end{subfigure}
    \hspace{0.02\linewidth} 
    \begin{subfigure}[b]{0.3\linewidth}
        \centering
        \includegraphics[width=\linewidth]{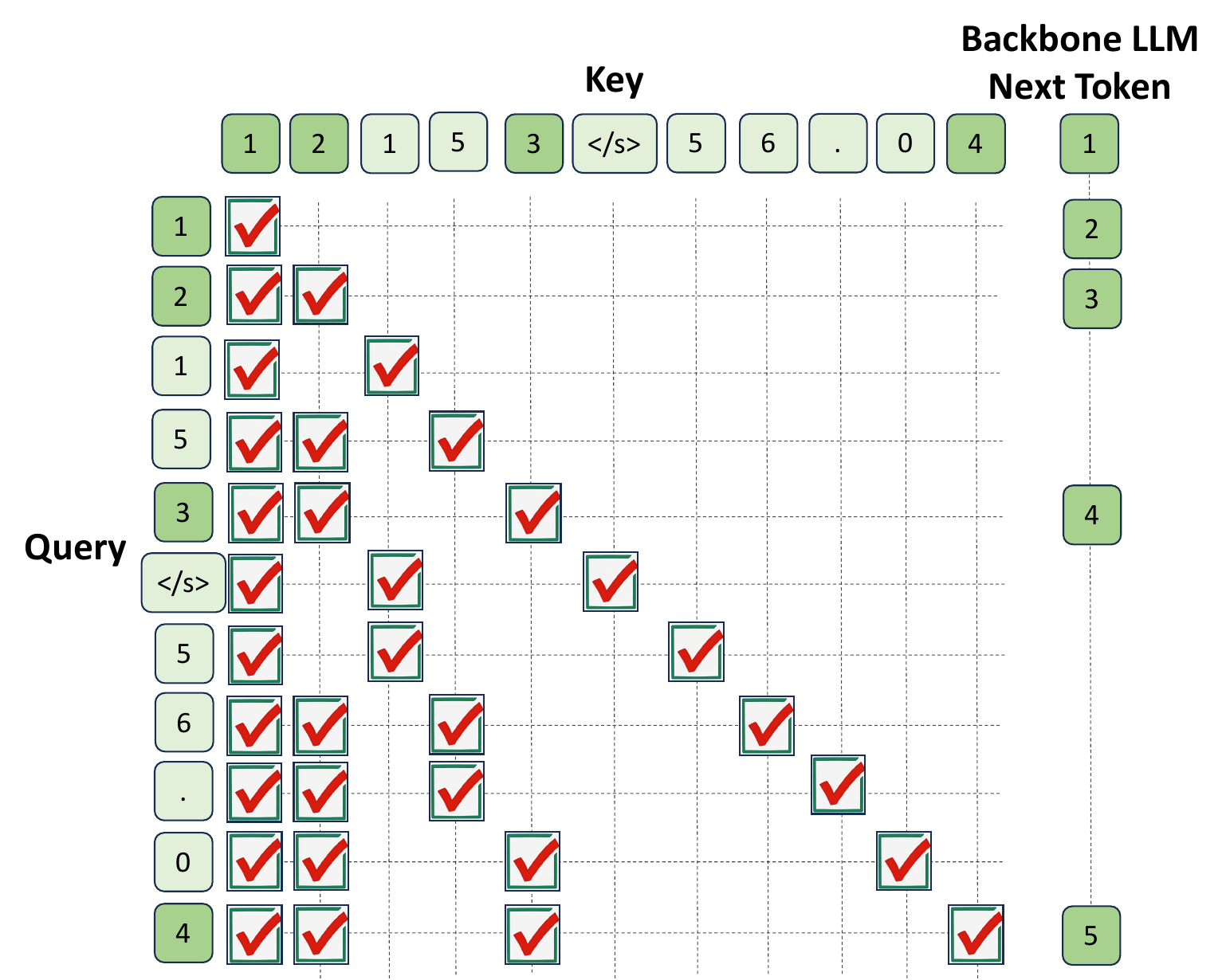}
        \caption{}
        \label{fig:attn_mask}
    \end{subfigure}
    \caption{(a) Architecture comparison between draft model of EAGLE and Scylla. We enhance Scylla with scaled decoders and output normalization (RMS Norm). (b) Specialized attention mask to efficiently validate candidate sequences.}
    \label{fig:combined_two_figures}
\end{figure}

\textbf{Training Paradigm.} To study the scaling laws of pretraining data and model capacity on speculative decoding, we implement the draft model through a two-stage training pipeline (Pretraining$\to$SFT) as illustrated in Figure~\ref{fig:architecture}. The Transformer-based EAGLE series architecture is chosen due to the extensively documented alignment of Transformer layers with potential scaling laws. We 
improve its architecture by incorporating embedding and output-layer normalization, while strategically scaling decoder capacity to investigate the scaling laws. To ensure controlled comparison validity, we maintain methodological parity with EAGLE2 through identical SFT data (shareGPT-68k), hyper-parameters, training loss configurations, and draft sampling procedures.
\begin{table}[ht]
  \centering
  \begin{minipage}[t]{0.40\textwidth}
    \centering
    \caption{pretraining data composition}
    \label{tab:pretraining-data}
    \begin{tabular}{lrr}
      \toprule
      Dataset & S.Prob & Count \\
      \midrule
      CC      & 80\%   & 93\,B  \\
      C4      & 15\%   & 9.5\,B \\
      GitHub  &  5\%   & 11\,B  \\
      \bottomrule
    \end{tabular}
  \end{minipage}\hspace{0.1em}%
  \begin{minipage}[t]{0.55\textwidth}
    \centering
    \caption{Iteration of Pretraining (PT) \& SFT}
    \label{tab:iteration}
    \begin{tabular}{lccc}
      \toprule
      Model    & PT    & SFT         & Total (/EAGLE2) \\
      \midrule
      EAGLE2   & –     & 86K         & $1\times$              \\
      EAGLE3   & –     & $8\times86$K & $8\times$             \\
      PT-10B   & 305K  & 86K         & $4.5\times$              \\
      \bottomrule
    \end{tabular}
  \end{minipage}
\end{table}

\textbf{Scaling Up pretraining Data.}
Following the data composition strategy of LLaMA~\cite{touvron2023llama}, we construct a heterogeneous pretraining corpus from Common Crawl (80\%), C4 (15\%), and GitHub (5\%) (see Table~\ref{tab:pretraining-data} and Appendix~\ref{app:pretrain_data}). We systematically scale pretraining token budgets across seven regimes (1B, 2B, 5B, 10B, 20B, 50B, 100B) to assess the impact of data scale on draft quality. Ablation experiments (Figure~\ref{fig:scaling_experiments} Left) demonstrate that increasing pretraining data from {1$\sim$100}\,B tokens increases from 5.13 to 5.43 (log-scaled x-axis) in acceptance rate. Notably, scaling pretraining data consistently outperforms scaling SFT data when compared to EAGLE3, and requires only 4.5$\times$ the training iterations to reach matching performance, versus EAGLE3’s 8$\times$ iterations (Table~\ref{tab:iteration}).

\textbf{Scaling Up Draft Model Capacity.}
Building on established LLM scaling laws, we isolate the decoder of the draft model for capacity expansion. For each backbone LLM, we create two larger draft variants by increasing decoder depth from 1,2 to 5 layers—corresponding to approximately 138M, 275M and 413M parameters. All other architectural hyperparameters (hidden dimension, number of attention heads, feed‑forward inner size) remain identical to the baseline.

\subsection{Roofline Model and Scaling Up Batch Size}
\label{sec:roofline model}

\textbf{Roofline Model for Speculative Decoding.}
To investigate strategies of scaling batch sizes for optimal decoding efficiency, we quantify computational workloads and memory access patterns based on Qwen2.5-72B model, whose architecture aligns with popular dense models (e.g. LLaMA family). Table~\ref{tab:analysis} summarizes key FLOPs and memory access components per decoding step, with complete derivations given in Appendix~\ref{app:analysis}.

\begin{table}[ht]
  \caption{Computational workload and memory access in a decoding step.}
  \label{tab:analysis}
  \centering
  \renewcommand{\arraystretch}{1.0} 
  \begin{tabular}{crrr}
    \toprule
    \textbf{OP}                 & \textbf{FLOPs}                    & \textbf{Read Access}                     & \textbf{Write Access}  \\
    \midrule
    \textbf{FC(2h$\to$1h)}                 & $4bsh^2$                          & $2bsh+2h^2$                              & $bsh$                  \\
    \midrule
    \textbf{QKV\_Proj}          & $2bsh(h+2\hkv)$                   & $bsh+(h+1)(h+2\hkv)$                     & $bs(h+2\hkv)$          \\
    \textbf{Self-Attention}     & $2 \times 2bs(s+\seqpre)h$        & $bsh+2b(s+\seqpre)\hkv$                  & $bsh$                  \\
    \textbf{Out\_Proj}          & $2bsh^2$                          & $bsh+h^2$                                & $bsh$                  \\
    \midrule
    \textbf{Up\&Gate\_Proj}     & $4bsh\hmlp$                       & $2(bsh+h\hmlp)$                          & $2bs\hmlp$             \\
    \textbf{Down\_Proj}         & $2bsh\hmlp$                       & $bs\hmlp+h\hmlp$                         & $bsh$                  \\
    \midrule
    \textbf{Residual}           & -                                 & $2\times 2bsh$                           & $2\times bsh$          \\
    \textbf{LayerNorm}          & -                                 & $2\times(h+bsh)$                         & $2\times bsh$          \\
    \textbf{LM\_Head}           & $2bshV$                           & $bsh+hV$                                 & $bsV$                  \\
    \bottomrule
  \end{tabular}
\end{table}

\begin{figure}[t]
    \centering
    \includegraphics[width=0.8\linewidth]{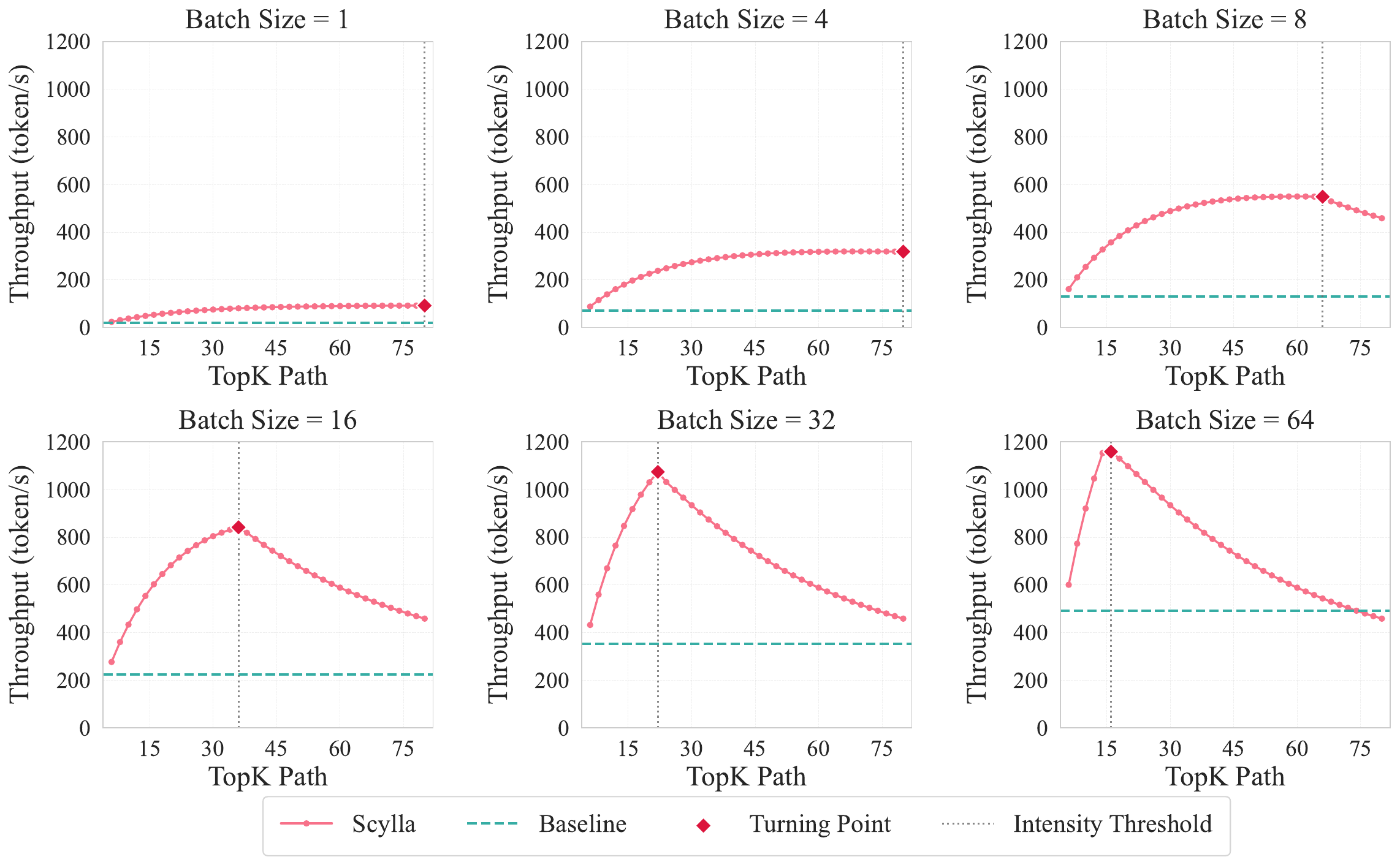}
    \caption{Theoretical throughput of Qwen2.5-72B (single-GPU) over batch size.}
    \label{fig:scaling_bs}
\end{figure}

In the decoding phase, the model's arithmetic intensity is defined as:

\begin{equation}
    \label{eq:arithmetic_intensity}
    \mathcal{I}(b, \topk) := \frac{\mathcal{W}_{\text{comp}}}{\mathcal{W}_{\text{mem}}}
\end{equation}

where $b$ denotes batch size, $\topk$ denotes TopK-path at verification stage, $\mathcal{W}_{\text{comp}}$ and $\mathcal{W}_{\text{mem}}$ represent computational workloads and memory access respectively. Although decoding stage is theoretically constrained by memory bandwidth, parallel evaluation of $\topk$ candidates induces arithmetic intensity growth proportional to batch size. This causes regime transitions from memory-bound ($\mathcal{I} < \mathcal{I}_{\text{crit}}$) to compute-bound ($\mathcal{I} \ge \mathcal{I}_{\text{crit}}$) operation. Its roofline can be described as: 

\begin{equation}
    \label{eq:I_crit}
    \mathcal{I}_{\text{crit}} := \frac{\mathcal{P}_{\text{peak}}}{\mathcal{B}_{\text{mem}}}
\end{equation}

where $\mathcal{P}_{\text{peak}}$ and $\mathcal{B}_{\text{mem}}$ denote peak performance and memory bandwidth respectively. This hardware-aware roofline model of speculative decoding enables systematic evaluation of operational proximity to hardware limitations across parameter configurations. Based on the roofline model, we present the relationship between theoretical throughput and $\topk$ under varying batch sizes as shown in Figure~\ref{fig:scaling_bs}. It is demonstrated that maximum throughput consistently occurs when the arithmetic intensity approaches $I_{\text{crit}}$.

\textbf{Optimal $\topk$ Configuration.} To maintain optimal throughput during batch size scaling, we derive the optimal $\topk$ criterion:

\begin{equation}
    \label{eq:optimal_topk}
    \mathcal{I}(b, \text{top}_{k, opt}) := \mathcal{I}_{\text{crit}}
\end{equation}

This formulation enables batch-adaptive optimal $\topk$ selection while avoiding performance degradation. We also analyze the interplay between throughput and acceptance rate in Appendix~\ref{app:interplay}. The empirical evidence conclusively demonstrates that the optimal $\topk$ follows a deterministic functional dependence on batch size ($b$), but independent with acceptance rate.

\section{Experiments} 
\label{sec:experients}

\textbf{Models and Tasks.}
We conduct extensive experiments on several widely-used SOTA open source language models including Vicuna-7B/13B~\cite{chiang2023vicuna}, Llama2-Chat-7B/13B/70B~\cite{touvron2023llama}, Qwen2.5-7B/72B~\cite{yang2024qwen2}, and Llama3-Instruct-8B/70B~\cite{grattafiori2024llama}.
For fair comparison with Medusa and EAGLE series, we use the identical evaluation protocols: MT-bench (MT~\cite{zheng2023judging}), HumanEval (HE~\cite{chen2021evaluating}), GSM8K~\cite{cobbe2021training}, Alpaca~\cite{taori2023alpaca}, CNN/Daily Mail (CD~\cite{nallapati2016abstractive}), and Natural Questions (NQ~\cite{kwiatkowski2019natural}). All experiments are conducted on NVIDIA H800 GPUs with single-GPU setup for 7B/8B/13B models and quad-GPU setup for 70B/72B models.

\textbf{Implementation.} The pretraining phase utilizes single-epoch training, while SFT stage employs two epochs with the ShareGPT-68K dataset, mirroring EAGLE2's configuration. In additional experiments following EAGLE3’s protocol, we scaled our SFT dataset by 8×—fine-tuning for two epochs on UltraChat-200K. We deliberately excluded EAGLE3’s hidden fusion and train-time test features to isolate the impact of data scaling and underscore the orthogonality of our work to EAGLE3’s innovations.

\textbf{Metrics.}
Our experiments follows two key constraints: frozen backbone LLM parameters and invariant acceptance criteria for lossless acceleration. This eliminates model performance evaluation, focusing instead on two key metrics: (1) \textit{Acceptance Rate}: Average accepted tokens per draft-verification cycle; (2) \textit{Throughput}: Tokens generated per second (token/s) during sustained decoding



\subsection{Scaling Up Pretraining Data}
\begin{figure}
\begin{center}
\includegraphics[width=0.8\linewidth]{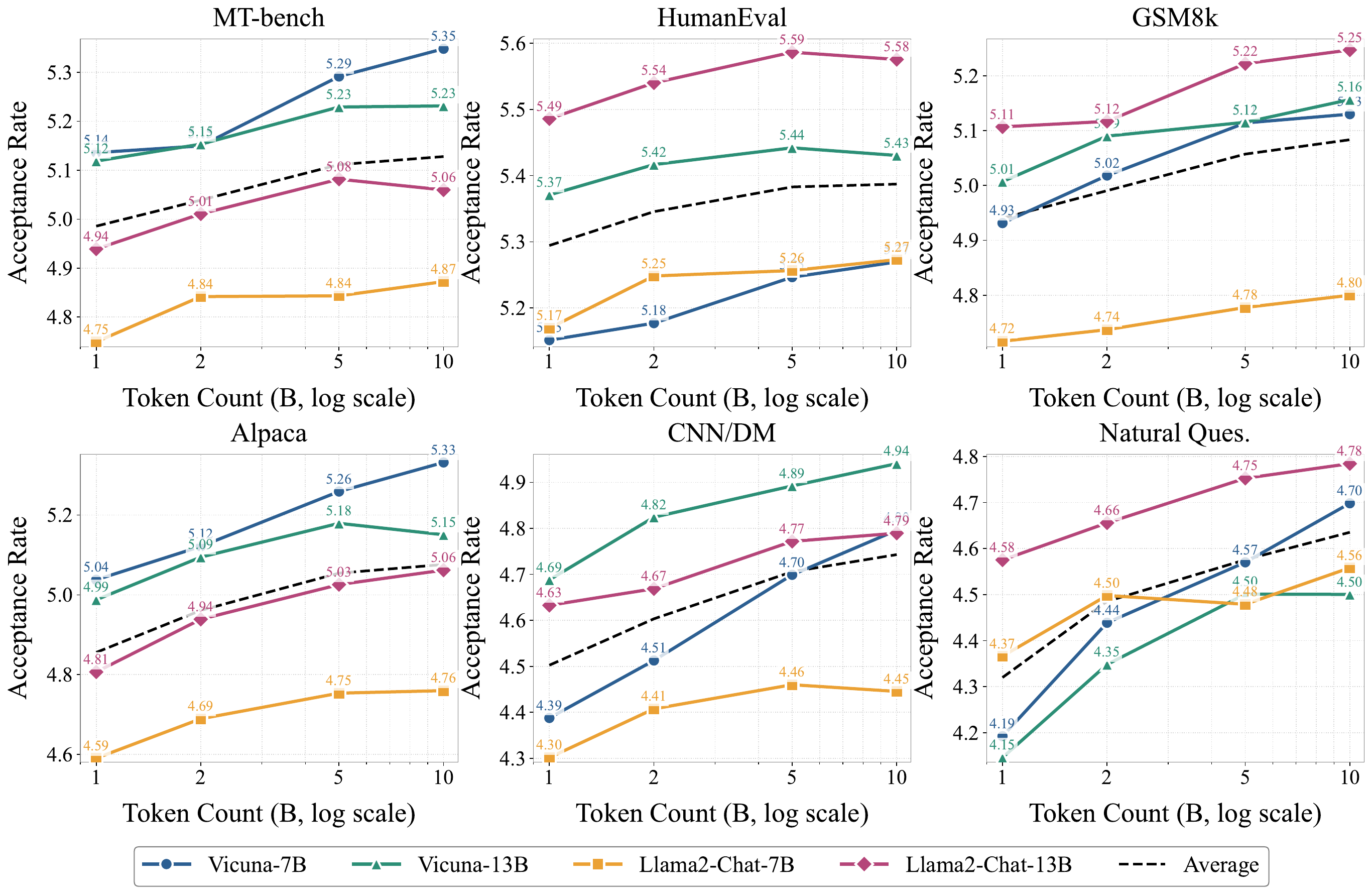}
\caption{Scaling up pretraining tokens.}
\label{fig:scaling_pt_token}
\end{center}
\end{figure}

In this section, we examine how scaling up pretraining tokens impacts acceptance rate. Figure~\ref{fig:scaling_experiments} Left panel shows acceptance rate exhibit logarithmic growth with pretraining tokens, increasing from 5.13 to 5.43 (log-scaled x-axis) compared to the constant baseline of 5.10 (without pretraining). However, acceptance rate gains saturate beyond 50 billion tokens, with minor fluctuations between 50 and 100 billion. These results validate that scaling pretraining data enhances performance but with diminishing returns at larger scales. The observed trend follows a log-linear scaling law (Theorem~\ref{theorem:scaling_law_pt_tokens}) with strong statistical validity ($R^2 = 0.89$), indicating a strong logarithmic correlation between acceptance rate and pretraining tokens. Scaling law of pretraining data also persists in other benchmarks shown in Figure~\ref{fig:scaling_pt_token}.

\subsection{Scaling Up Draft Model Capacity}
\begin{figure}
\begin{center}
\includegraphics[width=0.8\linewidth]{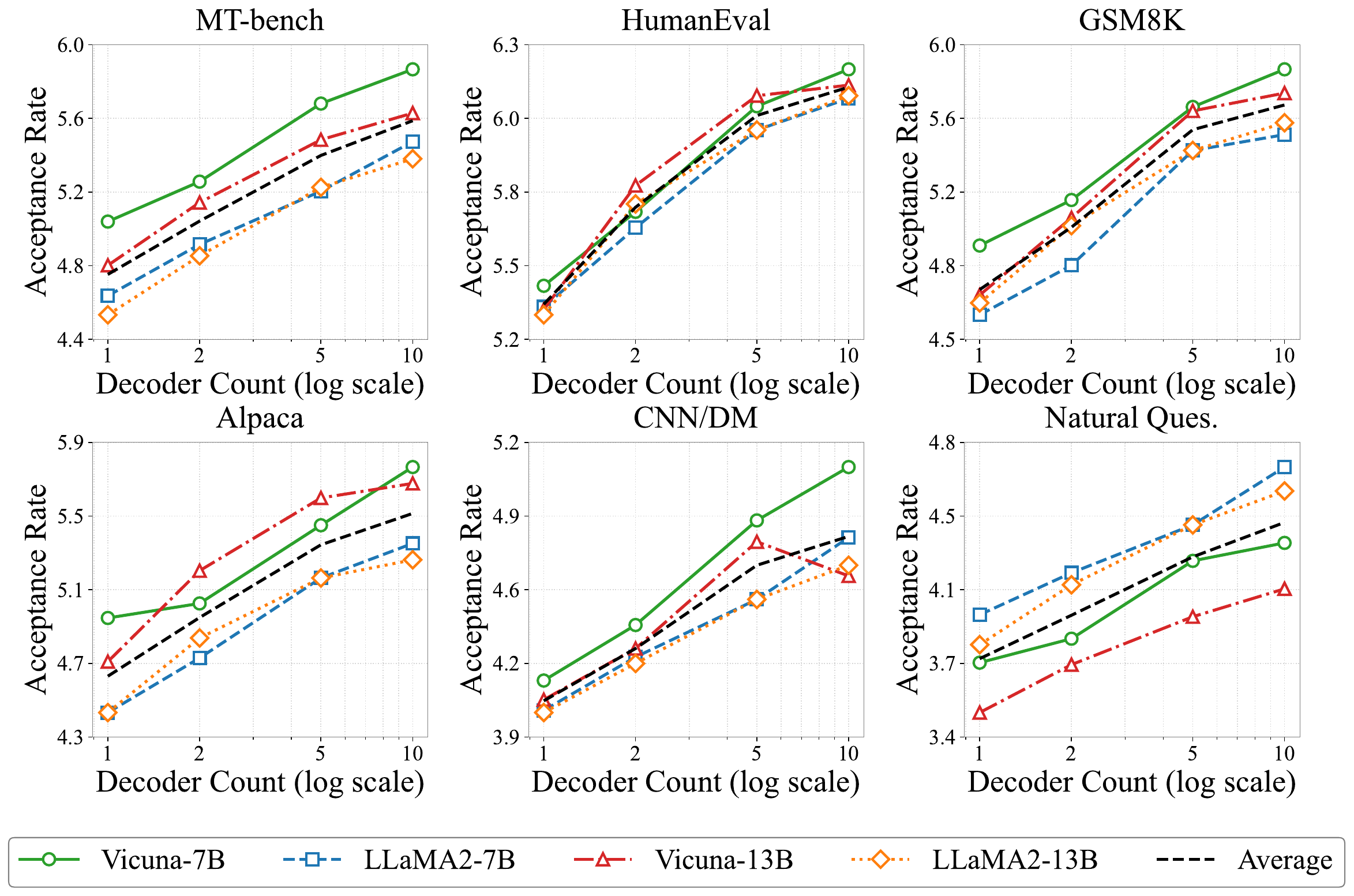}
\caption{Scaling up draft model capacity.}
\label{fig:scaling_decoder}
\end{center}
\end{figure}

Figure~\ref{fig:scaling_decoder} demonstrates a proportional improvement in acceptance rate across all models evaluated on six distinct benchmarks through draft decoder capacity scaling. The quasi-linear progression of average performance curve reveals the existence of scaling law governing the relationship between acceptance rate and draft decoder capacity (log scale). Notably, nearly all the benchmarks exhibit sustained performance gains up to 10 decoders, strongly indicating persistent optimization potential. This trend is also well captured by the log-linear formulation in Theorem~\ref{theorem:scaling_law_model_capacity}, confirming the mathematical relationship between acceptance rate and draft decoder capacity.

\subsection{Scaling Up Decoding Batch Size}
Our roofline analysis (Section~~\ref{sec:roofline model}) identifies peak throughput of speculative decoding at critical intensity threshold $\mathcal{I}_{\text{crit}}$ (Figure~\ref{fig:scaling_bs}). Suboptimal selection of $\topk$ induces performance degradation when $batch\ge8$, indicating the existence of decoding speed upper bound. By solving Equation~\ref{eq:optimal_topk} under Qwen2.5-72B architecture with a 10K prefill input, we formalize empirical scaling laws in Theorem~\ref{theorem:scaling_law_decoding_bs}. The right panel of Figure~\ref{fig:scaling_experiments} validates the inverse-batch square root scaling law between batch size ($b$) and TopK-path ($\topk$) fitted in Equation~\ref{eq:topk_vs_b}. The predicted optimal ($b$, $\topk$) configurations exhibit remarkable congruence with $\mathcal{I}_{\text{crit}}$ identified in Figure~\ref{fig:scaling_bs}. Furthermore, we establish a log-linear scaling law (Equation~\ref{eq:tps_vs_b}) for maximum throughput as a function of batch size. The regression validates Theorem~\ref{theorem:scaling_law_decoding_bs}'s efficacy in maximizing throughput during batch size scaling while avoiding throughput degradation.

\subsection{Scylla Built on Scaling Laws} 
Building on established scaling laws, we achieve Scylla through scaling pretraining with 10B tokens, draft model capacity and decoding batch size, which demonstrates superior performance across all evaluated models and tasks. As evidenced in Figure~\ref{fig1:best_result} and Table~\ref{table:results_restructured}, Scylla achieves substantial acceptance rate improvements of 28$\sim$53\% ($T=0$) and 24$\sim$42\% ($T=1$) over EAGLE2, with particularly significant gains on CD and NQ tasks. This enhancement stems from Scylla's pretraining-augmented knowledge acquisition, which extends beyond standard SFT to refine summarization and question-answering capabilities.

\begin{table}[t]
  \caption{Acceptance rates on various datasets for different models and methods ($T=0$ vs. $T=1$). The Medusa/EAGLE/EAGLE2 results are cited from their papers. See EAGLE3 in Appendix~\ref{app:eagle3}}
  \label{table:results_restructured}
  \centering
  \setlength{\tabcolsep}{2.65pt}  
  \renewcommand{\arraystretch}{1.0}  
  \begin{tabular}{cc *{7}{cc}}
    \toprule
    \multirow{2}{*}{\textbf{Model}} & \multirow{2}{*}{\textbf{Method}}
      & \multicolumn{2}{c}{\textbf{MT}}
      & \multicolumn{2}{c}{\textbf{HE}}
      & \multicolumn{2}{c}{\textbf{GSM8K}}
      & \multicolumn{2}{c}{\textbf{Alpaca}}
      & \multicolumn{2}{c}{\textbf{CD}}
      & \multicolumn{2}{c}{\textbf{NQ}}
      & \multicolumn{2}{c}{\textbf{Mean}} \\
    & & \textbf{T0} & \textbf{T1}
      & \textbf{T0} & \textbf{T1}
      & \textbf{T0} & \textbf{T1}
      & \textbf{T0} & \textbf{T1}
      & \textbf{T0} & \textbf{T1}
      & \textbf{T0} & \textbf{T1}
      & \textbf{T0} & \textbf{T1} \\
    \midrule
    \multirow{4}{*}{V7B}
      & Medusa   & 2.52 & –    & 2.67 & –    & 2.59 & –    & 2.48 & –    & 2.02 & –    & 2.09 & –    & 2.40 & –    \\
      & EAGLE    & 3.94 & 3.17 & 4.29 & 3.43 & 4.00 & 3.29 & 3.89 & 3.30 & 3.42 & 3.12 & 3.21 & 2.86 & 3.79 & 3.20 \\
      & EAGLE2   & 4.98 & 4.26 & 5.33 & 4.58 & 4.97 & 4.41 & 4.86 & 4.28 & 4.12 & 3.79 & 3.82 & 3.27 & 4.68 & 4.10 \\
      & Scylla   & \cellcolor{Col2}\textbf{6.45} & \cellcolor{Col2}\textbf{5.04} & \cellcolor{Col2}\textbf{6.62} & \cellcolor{Col2}\textbf{5.25} & \cellcolor{Col2}\textbf{6.29} & \cellcolor{Col2}\textbf{5.59} & \cellcolor{Col2}\textbf{6.40} & \cellcolor{Col2}\textbf{5.16} & \cellcolor{Col2}\textbf{6.34} & \cellcolor{Col2}\textbf{5.31} & \cellcolor{Col2}\textbf{5.79} & \cellcolor{Col2}\textbf{4.64} & \cellcolor{Col2}\textbf{6.32} & \cellcolor{Col2}\textbf{5.17} \\
    \cmidrule(lr){1-16}
    \multirow{3}{*}{LC7B}
      & EAGLE    & 3.62 & 3.30 & 4.24 & 3.79 & 3.82 & 3.52 & 3.71 & 3.33 & 3.41 & 3.15 & 3.44 & 3.12 & 3.71 & 3.37 \\
      & EAGLE2   & 4.70 & 4.44 & 5.39 & 4.96 & 4.77 & 4.57 & 4.66 & 4.37 & 4.12 & 3.83 & 4.19 & 3.88 & 4.64 & 4.34 \\
      & Scylla   & \cellcolor{Col2}\textbf{6.08} & \cellcolor{Col2}\textbf{5.44} & \cellcolor{Col2}\textbf{6.63} & \cellcolor{Col2}\textbf{5.81} & \cellcolor{Col2}\textbf{5.99} & \cellcolor{Col2}\textbf{5.42} & \cellcolor{Col2}\textbf{5.89} & \cellcolor{Col2}\textbf{5.35} & \cellcolor{Col2}\textbf{5.68} & \cellcolor{Col2}\textbf{5.25} & \cellcolor{Col2}\textbf{5.59} & \cellcolor{Col2}\textbf{5.20} & \cellcolor{Col2}\textbf{5.98} & \cellcolor{Col2}\textbf{5.41} \\
    \cmidrule(lr){1-16}
    \multirow{2}{*}{Q7B}
      & EAGLE2   & 4.06 & 3.60 & 4.69 & 4.49 & 4.33 & 4.14 & 4.43 & 4.20 & 3.54 & 3.09 & 3.26 & 2.91 & 4.05 & 3.74 \\
      & Scylla   & \cellcolor{Col2}\textbf{5.22} & \cellcolor{Col2}\textbf{4.52} & \cellcolor{Col2}\textbf{5.77} & \cellcolor{Col2}\textbf{5.34} & \cellcolor{Col2}\textbf{5.49} & \cellcolor{Col2}\textbf{5.33} & \cellcolor{Col2}\textbf{5.61} & \cellcolor{Col2}\textbf{5.10} & \cellcolor{Col2}\textbf{5.05} & \cellcolor{Col2}\textbf{4.16} & \cellcolor{Col2}\textbf{4.51} & \cellcolor{Col2}\textbf{3.99} & \cellcolor{Col2}\textbf{5.28} & \cellcolor{Col2}\textbf{4.74} \\
    \cmidrule(lr){1-16}
    \multirow{2}{*}{LI8B}
      & EAGLE2   & 4.19 & 3.89 & 4.78 & 4.62 & 4.36 & 4.16 & 4.62 & 4.19 & 3.78 & 3.46 & 3.36 & 3.20 & 4.18 & 3.92 \\
      & Scylla   & \cellcolor{Col2}\textbf{5.25} & \cellcolor{Col2}\textbf{4.86} & \cellcolor{Col2}\textbf{5.65} & \cellcolor{Col2}\textbf{5.43} & \cellcolor{Col2}\textbf{5.40} & \cellcolor{Col2}\textbf{5.16} & \cellcolor{Col2}\textbf{5.75} & \cellcolor{Col2}\textbf{5.21} & \cellcolor{Col2}\textbf{5.29} & \cellcolor{Col2}\textbf{4.74} & \cellcolor{Col2}\textbf{4.79} & \cellcolor{Col2}\textbf{4.43} & \cellcolor{Col2}\textbf{5.36} & \cellcolor{Col2}\textbf{4.97} \\
    \cmidrule(lr){1-16}
    \multirow{4}{*}{V13B}
      & Medusa   & 2.59 & –    & 2.78 & –    & 2.64 & –    & 2.45 & –    & 2.09 & –    & 2.10 & –    & 2.44 & –    \\
      & EAGLE    & 3.98 & 3.20 & 4.39 & 3.63 & 3.97 & 3.60 & 3.95 & 3.57 & 3.52 & 3.26 & 3.11 & 3.06 & 3.82 & 3.39 \\
      & EAGLE2   & 4.83 & 4.16 & 5.41 & 4.71 & 4.79 & 4.56 & 4.89 & 4.32 & 4.21 & 3.74 & 3.74 & 3.27 & 4.65 & 4.13 \\
      & Scylla   & \cellcolor{Col2}\textbf{6.37} & \cellcolor{Col2}\textbf{5.37} & \cellcolor{Col2}\textbf{6.90} & \cellcolor{Col2}\textbf{5.68} & \cellcolor{Col2}\textbf{6.26} & \cellcolor{Col2}\textbf{5.69} & \cellcolor{Col2}\textbf{6.34} & \cellcolor{Col2}\textbf{5.48} & \cellcolor{Col2}\textbf{6.41} & \cellcolor{Col2}\textbf{5.52} & \cellcolor{Col2}\textbf{5.56} & \cellcolor{Col2}\textbf{5.21} & \cellcolor{Col2}\textbf{6.31} & \cellcolor{Col2}\textbf{5.49} \\
    \cmidrule(lr){1-16}
    \multirow{3}{*}{LC13B}
      & EAGLE    & 3.90 & 3.45 & 4.52 & 3.78 & 4.03 & 3.67 & 3.83 & 3.55 & 3.59 & 3.39 & 3.47 & 3.31 & 3.89 & 3.53 \\
      & EAGLE2   & 4.75 & 4.36 & 5.52 & 5.17 & 4.90 & 4.62 & 4.61 & 4.26 & 4.24 & 3.84 & 4.04 & 3.75 & 4.68 & 4.33 \\
      & Scylla   & \cellcolor{Col2}\textbf{6.14} & \cellcolor{Col2}\textbf{5.64} & \cellcolor{Col2}\textbf{6.71} & \cellcolor{Col2}\textbf{6.08} & \cellcolor{Col2}\textbf{6.18} & \cellcolor{Col2}\textbf{5.71} & \cellcolor{Col2}\textbf{6.10} & \cellcolor{Col2}\textbf{5.62} & \cellcolor{Col2}\textbf{5.93} & \cellcolor{Col2}\textbf{5.55} & \cellcolor{Col2}\textbf{5.70} & \cellcolor{Col2}\textbf{5.37} & \cellcolor{Col2}\textbf{6.13} & \cellcolor{Col2}\textbf{5.66} \\
    \cmidrule(lr){1-16}
    \multirow{2}{*}{LC70B}
      & EAGLE2   & 4.26 & 4.26 & 5.02 & 5.04 & 4.42 & 4.31 & 4.07 & 4.11 & 3.67 & 3.57 & 3.57 & 3.56 & 4.17 & 4.14 \\
      & Scylla   & \cellcolor{Col2}\textbf{5.64} & \cellcolor{Col2}\textbf{5.54} & \cellcolor{Col2}\textbf{6.05} & \cellcolor{Col2}\textbf{6.01} & \cellcolor{Col2}\textbf{5.69} & \cellcolor{Col2}\textbf{5.65} & \cellcolor{Col2}\textbf{5.55} & \cellcolor{Col2}\textbf{5.48} & \cellcolor{Col2}\textbf{5.47} & \cellcolor{Col2}\textbf{5.36} & \cellcolor{Col2}\textbf{5.21} & \cellcolor{Col2}\textbf{5.20} & \cellcolor{Col2}\textbf{5.60} & \cellcolor{Col2}\textbf{5.54} \\
    \cmidrule(lr){1-16}
    \multirow{2}{*}{LI70B}
      & EAGLE2   & 4.23 & 4.12 & 5.12 & 4.98 & 4.50 & 4.45 & 4.81 & 4.75 & 3.60 & 3.54 & 3.51 & 3.44 & 4.68 & 4.33 \\
      & Scylla   & \cellcolor{Col2}\textbf{5.33} & \cellcolor{Col2}\textbf{5.20} & \cellcolor{Col2}\textbf{5.91} & \cellcolor{Col2}\textbf{5.82} & \cellcolor{Col2}\textbf{5.69} & \cellcolor{Col2}\textbf{5.66} & \cellcolor{Col2}\textbf{5.75} & \cellcolor{Col2}\textbf{5.76} & \cellcolor{Col2}\textbf{5.40} & \cellcolor{Col2}\textbf{5.25} & \cellcolor{Col2}\textbf{5.11} & \cellcolor{Col2}\textbf{5.01} & \cellcolor{Col2}\textbf{5.53} & \cellcolor{Col2}\textbf{5.45} \\
    \bottomrule
  \end{tabular}
\end{table}

 Theorem~\ref{theorem:scaling_law_decoding_bs} enables dynamic $\topk$ adjustment based on batch size. Experimental evaluation on our in-house inference engine (Table~\ref{table:decoing_peak_throughput}) reveals that while EAGLE2 outperforms the baseline at small batch size ($b \le 16$), its computational workloads escalates dramatically at larger batch sizes ($b \ge 32$), causing performance degradation. In contrast, Scylla + Opt.$\topk$ maintains superior throughput across all batch sizes through dynamic $\topk$ optimization guided by our scaling law. This strategy stabilizes arithmetic intensity near $I_{\text{crit}}$, preventing high computational overload during batch size scaling. Notably, Scylla + Opt.$\topk$ sustains a 1.21$\times$ speedup over baseline when $b=64$, outperforming EAGLE3 \cite{li2025eagle3} which ceases acceleration beyond $b>56$. Note that the observed throughput discrepancies between theory and practice arise from communication overhead and suboptimal kernel utilization, though they confirm our scaling law's fundamental validity. The eightfold SFT data scaling inspired by EAGLE3 yields extra 0.32 gains in acceptance rate over Scylla (see Appendix~\ref{app:eagle3} for details).

\begin{table}[ht]
  \caption{Decoding throughput (tokens/s) of Qwen2.5-72B (quad-GPU).}
  \label{table:decoing_peak_throughput}
  \centering
  \setlength{\tabcolsep}{2.65pt}  
  \renewcommand{\arraystretch}{0.9} 
  \begin{tabular}{crrrrrrrrr}
    \toprule
    \textbf{Batch size}             & 1 & 2 & 4 & 8 & 16 & 32 & 40 & 48 & 64 \\
    \midrule
    \textbf{Baseline w.o. SD}       & 60 & 118 & 224 & 428 & 750 & 1239 & 1406 & 1553 & 1763 \\
    \midrule
    \textbf{EAGLE2}                 & 200 & 400 & 650 & 810 & 830 & 910 & 1000 & 1010 & 980 \\
    \midrule
    \textbf{Scylla + Opt.$\topk$}      &\cellcolor{Col2}\textbf{250} & \cellcolor{Col2}\textbf{500} & \cellcolor{Col2}\textbf{800} & \cellcolor{Col2}\textbf{1200} & \cellcolor{Col2}\textbf{1650} & \cellcolor{Col2}\textbf{2050} & \cellcolor{Col2}\textbf{2100} & \cellcolor{Col2}\textbf{2200} & \cellcolor{Col2}\textbf{2150} \\
    \bottomrule
   \end{tabular}
\end{table}

\section{Related Works}
\label{sec:related_works}

Speculative decoding, first introduced as a blockwise parallel decoding framework by~\cite{stern2018blockwise}, accelerates auto-regressive generation through parallel token prediction. Subsequent refinements by~\cite{chen2023accelerating} and~\cite{leviathan2023fast} establish core methodologies for this field. Following these seminal works, the technique has gained significant attention. \cite{liu2023online}~addresses domain adaptation challenges through continuous online updates of draft models based on user queries. Structural innovations include Lookahead's tree-structured verification mechanism~\cite{zhao2024lookahead}, later optimized in Ouroboros through phrase candidate pooling~\cite{zhao2024ouroboros}, and further simplified by~\cite{fu2024break} through elimination of external data stores. Architectural modifications appear in Medusa's approach of merely integrating additional decoding heads on backbone LLMs~\cite{cai2024medusa}, subsequently enhanced through backbone model fine-tuning in Medusa2. Training methodology innovations include DistillSpec's knowledge distillation framework~\cite{zhou2023distillspec} for better draft-backbone alignment and CLLMs' optimization of Jacobi decoding efficiency~\cite{kou2024cllms}. System-level optimizations emerge in Sequoia's hardware-aware tree configuration algorithm~\cite{chen2024sequoia}, while \cite{marzollo2024sssd}~provides theoretical analysis highlighting batch size impacts on latency. Complementary approaches include \cite{gloeckle2024better}~and \cite{guo2025deepseek}'s demonstration of multi-token prediction benefits during pretraining. Input augmentation strategies surface in Clover series~\cite{xiao2024clover, xiao2024clover2}, while architectural refinements characterize EAGLE's draft model optimizations~\cite{li2024eagle, li2024eagle2, li2025eagle3} and HASS's train-time test procedure improvements~\cite{zhang2024learning}.

\section{Discussion and Limitations}
\label{sec:limitation}
While modern LLMs typically progress through three training stages: pretraining$\to$SFT$\to$RLHF, our investigation focuses solely on pretraining’s efficacy, leaving the potential benefits of RLHF for draft models unexplored. Although scaling pretraining data incurs no additional inference costs, enlarging draft model capacity introduces non-trivial computational overhead. A critical challenge lies in balancing draft model efficiency with acceptance rate, particularly in reducing the computational burden of its forward process without performance degradation. Our work primarily concentrates on dense LLM architectures, leaving draft model optimization for Mixture-of-Experts (MoE) LLMs as an open research direction. Furthermore, scaling laws in this work are verified only in Transformer-based Scylla, more experiments on other speculative decoding methods are left for the future.

\section{Conclusion}
In this work, we propose three empirical scaling laws of speculative decoding: (i) pretraining‐data scaling follows a log-linear law with diminishing returns beyond 50B tokens; (ii) draft-capacity scaling yields also log-linear improvements in acceptance rate; and (iii) decoding‐batch arithmetic intensity adheres to a roofline model that informs square root scaling law between batch size and optimal $\topk$. By jointly leveraging these laws, we achieve a 1.5-2.2 increase in acceptance rate and up to 2$\times$ higher decoding throughput over EAGLE2 under realistic large‐batch scenarios and 0.3 increase acceptance rate over EAGLE3. Future work will investigate RL fine‐tuning of the draft model to further boost acceptance rate.



\bibliographystyle{unsrt} 
\bibliography{custom}

\appendix

\newpage

\section{Pretraining Data Selection Details}
\label{app:pretrain_data}
The pretraining data used in the experiments are achieved from the Hugging-Face.
\begin{itemize}
  \item Common Crawl: togethercomputer/RedPajama-Data-1T, first 60 shards
  \item GitHub: togethercomputer/RedPajama-Data-1T, first 25 shards
  \item C4: allenai/c4, first 50 JSON files
\end{itemize}

\section{Experiments Details}
All our experiments use dense models; although other acceleration techniques such as quantization and distillation~\cite{sinkd,multi,sink2} are available, we adopt speculative decoding for its ability to achieve lossless speedup~\cite{liu2024kangaroo}.

\subsection{Models}
To ensure reproducibility, our experiments employ several publicly available language models:  Vicuna-7B/13B~\cite{chiang2023vicuna}, Llama2-Chat-7B/13B/70B~\cite{touvron2023llama}, Qwen2.5-7B/72B~\cite{yang2024qwen2}, and Llama3-Instruct-8B/70B~\cite{grattafiori2024llama}. We outline their technical specifications below:

\paragraph{Vicuna}
Vicuna~\cite{chiang2023vicuna} is an open-source LLM developed by researchers from UC Berkeley, CMU, Stanford, UC San Diego, and MBZUAI, which adapts LLaMA foundation model using approximately 70,000 user-shared conversational data collected from ShareGPT. The training process incorporates several optimizations, including extending the maximum context length from 512 to 2048 tokens to enhance long-sequence understanding, leveraging gradient checkpointing and flash attention for memory efficiency, and refining loss computation to prioritize multi-turn dialogue coherence. Preliminary results indicate that Vicuna-13B achieves 92\% of ChatGPT’s performance in aggregate scores, outperforming LLaMA and Alpaca in over 90\% of cases and matching or surpassing ChatGPT in 45\% of scenarios. 

\paragraph{LLaMA 2}
Meta's Llama 2 series~\cite{touvron2023llama} introduces three model sizes (7B/13B/70B) trained on 2 trillion tokens from non-proprietary sources. The architecture builds upon the standard transformer framework with three-stage optimization: pretraining, SFT, and human preference alignment through Proximal Policy Optimization (PPO). Notable enhancements over the original LLaMA include a 4k-token context window, strengthened safety guardrails, and improved performance on reasoning tasks (commonsense, coding, and knowledge-intensive benchmarks).

\paragraph{LLaMA 3}
Llama 3~\cite{grattafiori2024llama}, also developed by Meta, introduces 8B/70B Transformer models with critical innovations such as Group Query Attention (GQA) to enhance inference efficiency and Rotary Positional Encoding (RoPE) for robust long-sequence processing, supporting a context window of 8,192 tokens. The model employs a 128K-token vocabulary, significantly improving text encoding efficiency compared to Llama 2.

\paragraph{Qwen}
Qwen2.5~\cite{yang2024qwen2} demonstrates significant improvements in reasoning capabilities and contextual understanding over its predecessors, achieved through scaled model parameters (ranging from 0.5B to 72B variants) and optimized attention mechanisms. The training corpus incorporates 3 trillion filtered multilingual tokens spanning technical literature, web content, and conversational datasets. Notable enhancements include an extended context window supporting 128k token sequences through RoPE optimizations, coupled with improved training stability via dynamic gradient clipping strategies. Quantitative evaluations on standardized benchmarks (MMLU, GSM8K, HumanEval) demonstrate 15-20\% absolute performance gains compared to Qwen2, particularly in STEM-related problem solving and code generation tasks.

\subsection{Benchmark Selection Details}
\label{app:benchmark_data}
For fair comparison with Medusa and EAGLE series, we use the identical evaluation protocols: MT-bench (MT~\cite{zheng2023judging}) for multi-turn conversation assessment, HumanEval (HE~\cite{chen2021evaluating}) for code generation, GSM8K~\cite{cobbe2021training} for mathematical reasoning, Alpaca~\cite{taori2023alpaca} for instruction following, CNN/Daily Mail (CD~\cite{nallapati2016abstractive}) for summarization, and Natural Questions (NQ~\cite{kwiatkowski2019natural}) for question answering. All experiments are conducted on NVIDIA H800 GPUs with single-GPU configuration for 7B/8B/13B models and quad-GPU setup for 70B/72B models.

\section{Computational Workloads and Memory Access Analysis}
\label{app:analysis}

\subsection{Mathematical Notations}

\begin{table}[ht]
  \caption{Mathematical Notations.}
  \label{table:notation}
  \centering
  \renewcommand{\arraystretch}{1.1} 
  \begin{tabular}{rl}
    \toprule
    \textbf{Symbol} & \textbf{Description}.           \\
    \midrule
    $b$       & Batch size                            \\
    $s$       & Decoding sequence length              \\
    $\seqpre$ & Prefilled sequence length             \\
    $h$       & Hidden dimension                      \\
    $\hkv$    & Hidden dimension of Key and Value     \\
    $\hmlp$   & Hidden dimension of MLP               \\
    $n_h$     & Number of attention heads             \\
    $V$       & Vocabulary size                       \\
    $l$       & Layers of target model                \\
    $D$       & Depth of decoding draft model         \\
    $\tacc$   & Accepted tokens of target model       \\
    $k$       & TopK tokens generated by draft model  \\
    $\topk$   & TopK path of draft model              \\
    \bottomrule
  \end{tabular}
\end{table}

\subsection{Computational Workload}

\begin{table}[ht]
  \caption{Computational Workloads in a Scylla Forward Pass.}
  \label{table:computational_workloads}
  \centering
  \renewcommand{\arraystretch}{1.2} 
  \begin{tabular}{ccc}
    \toprule
    \textbf{Component}                      & \textbf{OP}              & \textbf{FLOPs}            \\
    \midrule
    \multirow{6}{*}{\textbf{Transformer}}   & \textbf{Q, K, V}         & $2bsh(h+2\hkv)$  \\
                                            & \textbf{Self Attention}  & $2\times 2bs(s+\seqpre)h$ \\
                                            & \textbf{Output Proj}     & $2bsh^2$                  \\
                                            & \textbf{MLP}             & $6bsh\hmlp$      \\
                                            & \textbf{LM Head}         & $2bsV$                    \\
    \midrule
    \textbf{Draft Model}                    & \textbf{FC}              & $4bsh^2$                  \\
    \bottomrule
  \end{tabular}
\end{table}

The computation of a Scylla forward pass is made up of three parts: target model (transformer), decoding draft model and prefilling draft model.

\begin{itemize}
    \item Transformer: For $l$ decoder layers in a forward pass with $\topk$ input tokens, the total computation is: $2b\times \topk \times h[V + l(2h + 2\topk + 2\seqpre + 2\hkv + 3\hmlp)]$ FLOPs.
    
    \item Decoding draft model
    \begin{itemize}
        \item FC: GEMM($[s, 2h][2h, h]$), accounting for $4bsh^2$ FLOPs.
        \item Rest computational analysis is the same as above.
    \end{itemize}
    For $D$ decoding draft model with $k$ input tokens ($s=k$), the total computation is : $D \times [2bkhV + 4bkh^2 +  (4bkh^2+4bk(k + \seqpre)h+4bkh\hkv + 6bkh\hmlp)] = D \times 2bkh(V + 4h + 2k + 2\seqpre + 2\hkv + 3\hmlp)$ FLOPs.
    
    \item Prefilling draft model \\
    The computational analysis is the same as above. Therefore, for one prefilling draft model with $\tacc$ input tokens ($s=\tacc$), the total computation is: $2b \times \tacc \times h(V + 4h + 2 \times \tacc +2\seqpre +2\hkv +3\hmlp)$ FLOPs.
\end{itemize}

In summary, the total computation of a Scylla forward pass is: $2bsh[V + l(2h + 2s + 2\seqpre + 2\hkv + 3\hmlp)] + D \times 2bkh(V + 4h + 2k + 2\seqpre + 2\hkv + 3\hmlp) + 2b \times \tacc \times h(V + 4h + 2 \times \tacc +2\seqpre +2\hkv +3\hmlp)$ FLOPs.

\subsection{Memory Access}

\begin{table}[ht]
  \caption{Memory Access in a Scylla Forward Pass.}
  \label{table:memory_access }
  \centering
  \renewcommand{\arraystretch}{1.2} 
  \begin{tabular}{ccrr}
    \toprule
    \textbf{Component}                      & \textbf{OP}                       & \textbf{Read Access}     & \textbf{Write Access} \\
    \midrule
    \multirow{8}{*}{\textbf{Transformer}}   & \textbf{Residual $\times 2$}      & $2\times 2bsh$           & $2\times bsh$         \\
                                            & \textbf{LayerNorm $\times 2$}     & $2\times (h+bsh)$        & $2\times bsh$         \\
                                            & \textbf{QKV with Bias}            & $bsh+(h+1)(h+2\hkv)$     & $bs(h+2\hkv)$         \\
                                            & \textbf{Self Attention}           & $bsh+2b(s+\seqpre)\hkv$  & $bsh$                 \\
                                            & \textbf{Output Proj}              & $bsh+h^2$                & $bsh$                 \\
                                            & \textbf{MLP}                      & $2bsh+3h\hmlp+bs\hmlp$   & $2bs\hmlp+bsh$        \\
                                            & \textbf{Activation}               & $2bs\hmlp$               & $bs\hmlp$             \\
                                            & \textbf{LM Head}                  & $bsh+hV$                 & $bsV$                 \\
    \midrule
    \textbf{Draft Model}                    & \textbf{FC}                       & $2bsh+2h^2$              & $bsh$                 \\
    \bottomrule
  \end{tabular}
\end{table}

The memory access of a Scylla forward pass is made up of three parts: target model (transformer), decoding draft model and prefilling draft model.

\begin{itemize}
    \item Transformer: For $l$ decoder layers with $\topk$ input tokens, the total memory access is: 
    \begin{itemize}
        \item Read access: $l[11b\times \topk\times h + 2b(\topk+\seqpre)\hkv +3b\times \topk\times \hmlp+ 2h^2 + 2h\cdot \hkv + 3h\cdot \hmlp +3h+2\hkv] + b\times \topk\times h + hV$
        \item Write access: $l(8b\times \topk\times h + 2b\times \topk\times \hkv + 3b\times \topk\times \hmlp) + b\times \topk\times V$
    \end{itemize}
    
    \item Decoding draft model: For $D$ decoding draft model with $k$ input tokens ($s=k$), the total memory access is: 
    \begin{itemize}
        \item Read access: $D \times [13bkh+ 2b(k+\seqpre)\hkv +3bk\hmlp+ 4h^2 + 2h\cdot \hkv + 3h\cdot \hmlp +3h+2\hkv + bkh + hV]$
        \item Write access: $D \times (9bkh + 2bk\hkv + 3bk\hmlp + bkV)$
    \end{itemize}
    
    \item Prefilling draft model: For one prefilling draft model with $\tacc$ input tokens, the total memory access is: 
    \begin{itemize}
        \item Read access: $13b \times \tacc \times h+ 2b(\tacc+\seqpre)\hkv +3b \times \tacc \times \hmlp+ 4h^2 + 2h\cdot \hkv + 3h\cdot \hmlp +3h+2\hkv + b \times \tacc \times h + hV $
        \item Write access: $b \times \tacc \times (9h + 2\hkv + 3\hmlp +  V) $
    \end{itemize}
\end{itemize}

In summary, the total memory of a Scylla forward pass is:
\begin{itemize}
    \item Read access: $13b \times \tacc \times h+ 2b(\tacc+\seqpre)\hkv +3b \times \tacc \times \hmlp+ 4h^2 + 2h\cdot \hkv + 3h\cdot \hmlp +3h+2\hkv + b \times \tacc \times h + hV + 5 \times [13b \times k \times h+ 2b(k+\seqpre)\hkv +3b \times k \times \hmlp+ 4h^2 + 2h\cdot \hkv + 3h\cdot \hmlp +3h+2\hkv + b \times k \times h + hV] + l[11b \times (top_k + 1) \times h + 2b(top_k + 1 +\seqpre)\hkv +3b \times (top_k + 1) \times \hmlp+ 2h^2 + 2h\cdot \hkv + 3h\cdot \hmlp +3h+2\hkv] + b \times (top_k + 1) \times h + hV$
    \item Write access: $b \times \tacc \times (9h + 2\hkv + 3\hmlp +  V) + D \times b \times k \times  (9h + 2\hkv + 3\hmlp +  V) + b \times (top_k + 1)[l(8h + 2\hkv + 3\hmlp)+ V]$
\end{itemize}

\section{Additional Experiments on EAGLE3}
\label{app:eagle3}
We extend our SFT setup to match EAGLE3’s data-scaling protocol by applying an 8× augmentation of the UltraChat-200K corpus and running two epochs of fine-tuning on eight H800 GPUs over approximately eight hours. Note that, under EAGLE3’s original configuration, the SFT stage spans 20 epochs and takes roughly 80 hours. Under this controlled, data-only scaling regime—excluding EAGLE3’s hidden fusion and train-time test innovations—we observe an absolute acceptance rate increase of 0.3$\sim$0.5 over the Scylla baseline and 0.32 over EAGLE3 (on Vicuna 13B). These results confirm that sheer data volume alone drives substantial gains in Transformer-based draft models.

\begin{table}[t]
  \caption{Acceptance rates on various datasets for different models and methods ($T=0$).}
  \label{table:eagle3_results}
  \centering
  \renewcommand{\arraystretch}{1.0}  
  \begin{tabular}{cc *{7}{c}}
    \toprule
    \textbf{Model} & \textbf{Method}
      & \textbf{MT} & \textbf{HE}
      & \textbf{GSM8K} & \textbf{Alpaca}
      & \textbf{CD} & \textbf{NQ}
      & \textbf{Mean} \\
    \midrule
    \multirow{3}{*}{V7B}
      & EAGLE       & 3.94 & 4.29 & 4.00 & 3.89 & 3.42 & 3.21 & 3.79 \\
      & Scylla      & 6.45 & 6.62 & 6.29 & 6.40 & 6.34 & 5.79 & 6.32 \\
      & Scylla+8SFT & \cellcolor{Col2}\textbf{6.76} & \cellcolor{Col2}\textbf{6.88} & \cellcolor{Col2}\textbf{6.42} & \cellcolor{Col2}\textbf{6.81} & \cellcolor{Col2}\textbf{6.90} & \cellcolor{Col2}\textbf{6.26} & \cellcolor{Col2}\textbf{6.67} \\
    \cmidrule(lr){1-9}
    \multirow{3}{*}{V13B}
      & EAGLE       & 3.98 & 4.39 & 3.97 & 3.95 & 3.52 & 3.11 & 3.82 \\
      & EAGLE3      & 6.65 & 7.54 & 6.29 & 6.17 & 6.47 & -    & 6.62 \\
      & Scylla      & 6.37 & 6.90 & 6.26 & 6.34 & 6.41 & 5.56 & 6.31 \\
      & Scylla+8SFT & \cellcolor{Col2}\textbf{6.86} & \cellcolor{Col2}\textbf{7.25} & \cellcolor{Col2}\textbf{6.51} & \cellcolor{Col2}\textbf{7.06} & \cellcolor{Col2}\textbf{7.03} & \cellcolor{Col2}\textbf{6.15} & \cellcolor{Col2}\textbf{6.81} \\
    \cmidrule(lr){1-9}
    \multirow{3}{*}{LC7B}
      & EAGLE       & 3.62 & 4.24 & 3.82 & 3.71 & 3.41 & 3.44 & 3.71 \\
      & Scylla      & 6.08 & 6.63 & 5.99 & 5.89 & 5.68 & 5.59 & 5.98 \\
      & Scylla+8SFT & \cellcolor{Col2}\textbf{6.34} & \cellcolor{Col2}\textbf{6.78} & \cellcolor{Col2}\textbf{6.19} & \cellcolor{Col2}\textbf{6.36} & \cellcolor{Col2}\textbf{6.17} & \cellcolor{Col2}\textbf{6.11} & \cellcolor{Col2}\textbf{6.33} \\
    \cmidrule(lr){1-9}
    \multirow{3}{*}{LC13B}
      & EAGLE       & 3.90 & 4.52 & 4.03 & 3.83 & 3.59 & 3.47 & 3.89 \\
      & Scylla      & 6.14 & 6.71 & 6.18 & 6.10 & 5.93 & 5.70 & 6.13 \\
      & Scylla+8SFT & \cellcolor{Col2}\textbf{6.40} & \cellcolor{Col2}\textbf{7.10} & \cellcolor{Col2}\textbf{6.29} & \cellcolor{Col2}\textbf{6.51} & \cellcolor{Col2}\textbf{6.29} & \cellcolor{Col2}\textbf{6.00} & \cellcolor{Col2}\textbf{6.43} \\
    \bottomrule
  \end{tabular}
\end{table}

\section{Interplay Between Acceptance Rate and Throughput}
\label{app:interplay}

To systematically investigate the influence of acceptance rate on throughput, we derive an empirical relationship between acceptance rate ($\tacc$) and TopK-path ($\topk$) through experimental analysis. The established functional relationship can be expressed as below:

\begin{equation}
    \label{eq:acc_vs_topk}
    \tacc = -6.25 \times (0.2k)^{\frac{\topk}{30}} + 6
\end{equation}

where $k$ denotes an experimentally determined scaling factor ranging from 0.9 to 1.2. This parametric formulation captures the nonlinear coupling between acceptance rate and TopK-path selection observed in our experimental configurations.

\begin{figure}[ht]
    \centering
    \includegraphics[width=0.9\linewidth]{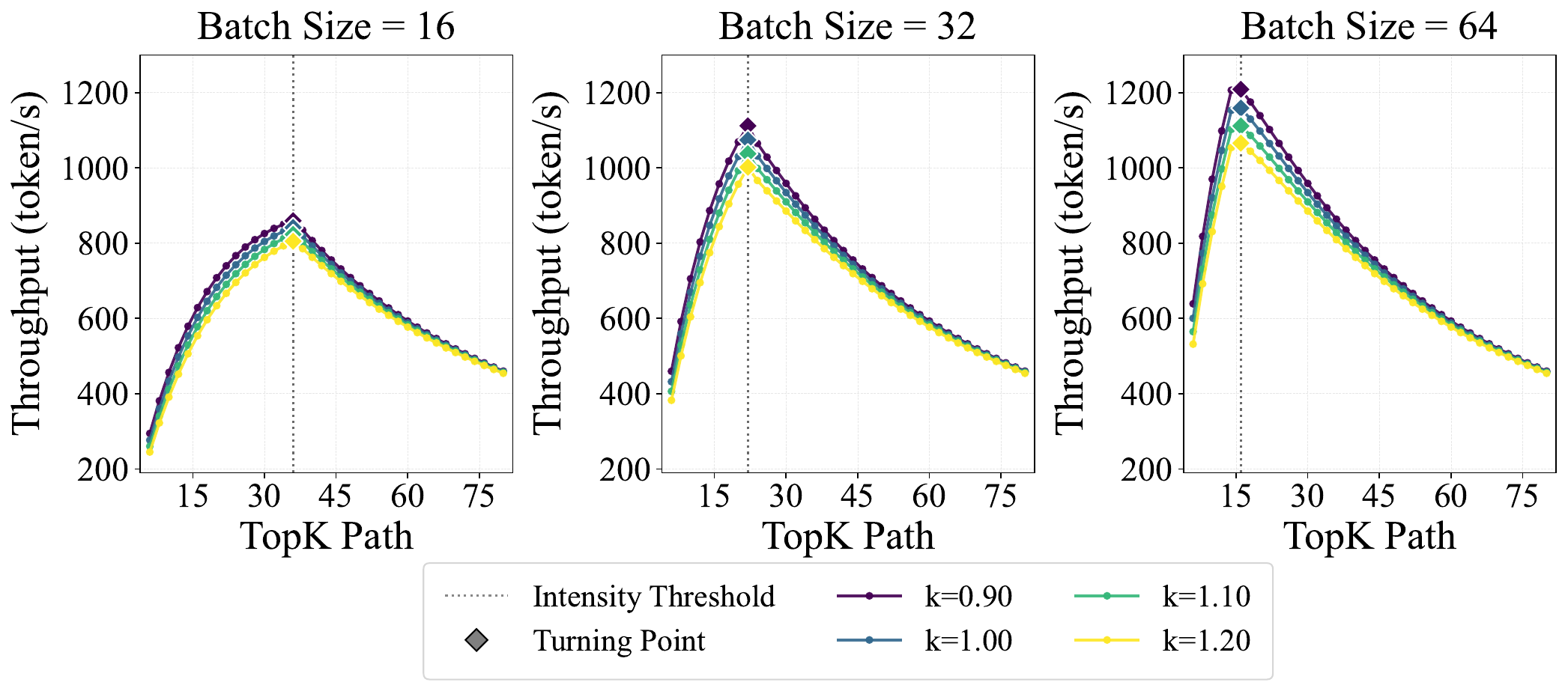}
    \caption{Throughput variations across different $\topk$ values while accounting for acceptance rate fluctuations on batch sizes of 16, 32 and 64.}
    \label{fig:acc_analysis}
\end{figure}

Using this formula, we systematically evaluate throughput variations across different $\topk$ values while accounting for acceptance rate fluctuations on batch sizes of 16, 32 and 64. The results are shown in the Figure~\ref{fig:acc_analysis}, revealing two critical obeservations: (1) Maximum throughput consistently occurs near $I_{crit}$, and (2) The optimal $\topk$ demonstrates negligible dependence on acceptance rate variations. These findings collectively suggest that while acceptance rate modulates absolute throughput values, $\topk$ operates primarily as an independent control variable.

\end{document}